\documentclass[a4paper,11pt,openright,twoside]{article}
\usepackage[english,francais]{babel}
\usepackage[latin1]{inputenc}
\usepackage[T1]{fontenc}

\usepackage{vmargin}
\usepackage{pstricks,pst-node,pst-text,pst-3d}
\usepackage[normalem]{ulem}

\def\nom{Jacquemin}
\def\prenom{Bernard }
\def\auteur{\prenom \textsc{\nom}}

\def\adresselabo{Institut des Sciences Cognitives\\
                 CNRS -- UMR5015\\
                 67, Bd Pinel\\
		 69675 Bron cedex (France)}

\def\courriel{Bernard.Jacquemin@isc.cnrs.fr}

\def\titre{Interroger un corpus par le sens\\Une approche linguistique}

\def\titrecourt{Interroger un corpus par le sens}

\def\piedpage{\textit{«~Mots, termes, contextes~». Actes des 7\up{e} Journées Scientifiques du réseau LTT}\\Bruxelles, 8-10 septembre 2005, pp. 347-360.}

\usepackage{fancyhdr}
\title{\titre}
\author{\auteur\\\adresselabo\\\courriel}
\date{}

\begin{document}
\thispagestyle{empty}
\maketitle

\begin{abstract}
\noindent Constatant que les méthodes statistiques dominantes en traitement de l'information ne
peuvent résoudre certaines difficultés, je propose une approche centrée sur les méthodes
linguistiques symboliques afin d'identifier la contribution que ces dernières peuvent apporter
au domaine. Elle s'appuie sur une identification du sens des mots et des relations
entre ces mots pour proposer des reformulations d'énoncés sans changement de signification.
Les reformulations sont parasynonymiques et dérivationnelles et permettent
de trouver une information textuelle quelle que soit la formulation de l'information ou
de la requête. \\
\textbf{Mots-clefs}: Sémantique lexicale; désambiguïsation; question-réponse; extraction d'information; corpus ; génération; reformulation; analyse syntaxico-sémantique; synonymie; dérivation morphologique; dictionnaire électronique
\end{abstract}

\selectlanguage{english}
\begin{abstract}
\noindent In textual knowledge management, statistical methods prevail. Nonetheless, some difficulties
cannot be overcome by these methodologies. I propose a symbolic approach
using a complete textual analysis to identify which analysis level can improve the
the answers provided by a system. The approach identifies word senses and relation
between words and generates as many rephrasings as possible. Using synonyms and
derivative, the system provides new utterances without changing the original meaning
of the sentences. Such a way, an information can be retrieved whatever the question
or answer's wording may be. 
\end{abstract}
\selectlanguage{francais}

\section{Introduction}

La société actuelle a fait de la maîtrise de l'information un enjeu de savoir autant que de pouvoir. Cependant, face à la profusion des sources d'information, face à l'enchevêtrement ingérable des données elles-mêmes, personne n'est plus capable de fournir un accès rapide à un élément d'information précis. Les initiatives qui visent à élaborer une méthode automatique de gestion de l'information capable d'ordonner des masses de données 
sont dès lors bienvenues.

Dans les champs de recherche liés à l'information textuelle électronique, et notamment la tâche de question-réponse, plusieurs méthodes ont vu le jour, qui permettent de confronter les données de la question avec celles contenues dans un texte. Si les données correspondent, on considère que la réponse à la question posée est dans le contexte immédiat de l'information commune à la question et à la bribe de texte. Diverses conférences internationales ont également vu le jour, dont l'objet est l'évaluation des systèmes proposés: TREC, CLEF, NTCIR\dots

Il reste toutefois que toutes ces méthodes fonctionnent sur base d'un même schéma. Il s'agit en effet d'appréhender la question, de l'analyser pour la débarrasser de tout élément perturbateur, et d'en effectuer une expansion destinée à contrer les variations de forme qui peuvent se présenter dans les documents interrogés. De plus, les meilleurs systèmes plaident tous pour des approches capable de gérer au mieux le caractère langagier des textes. Par exemple, \cite{Hull99} proposait déjà l'exploitation des résultats d'une analyse morphologique pour indexer les éléments significatifs tant dans les requêtes que dans les réponses. Par la suite, \cite{FerretAl02a} ont proposé une certaine intégration de la syntaxe, capable de reconnaître un certain nombre d'entités nommées ainsi que de déterminer la nature de la réponse à la question. Le meilleur système actuellement disponible \cite{HarabagiuAl02} s'appuie sur des notions sémantiques issues du réseau WordNet, ainsi que sur un moteur d'inférences logiques pour proposer un niveau très élevé de réponses correctes. 

De plus, une consultation, même rapide, des publications liées aux campagnes d'évaluation du domaine de question-réponse \cite{VoorheesBuckland04,PetersAl04,NTCIR4} permet de constater que les qualités qui distinguent le résultat des différentes approches résident dans leur capacité à mieux appréhender la langue, à obtenir une meilleure analyse linguistique non seulement de la question, mais également des réponses possibles. Malgré les vertus reconnues des modules d'analyse linguistique, il est pourtant étrange de constater qu'ils n'occupent qu'une place générique dans toutes ces approches à dominante statistique, et qu'aucune recherche n'est actuellement menée pour leur accorder un statut plus central qui pourrait encore améliorer le fonctionnement des logiciels.

Cet article s'appuie sur les constats précédents pour proposer une méthode où l'analyse linguistique occupe une place prépondérante à tous les niveaux du système. Nous allons d'abord présenter les contraintes propres à une analyse morpho-syntaxique et sémantique d'énoncés textuels et exposer les choix auxquels elles nous ont amené. Ensuite, nous présenterons brièvement les outils d'analyse que nous avons utilisés dans notre approche, ainsi que les ressources lexico-sémantiques que nous avons exploitées, et les adaptations qui ont été nécessaires. Après cela, nous présenterons la construction d'une structure informationnelle qui fournit un accès à chaque élément d'information contenu dans une base textuelle. Enfin, nous présenterons la méthode permettant de trouver la réponse à une question posée en français dans cette base textuelle. Finalement, nous présenterons quelques perspectives futures pour ce sujet de recherche, et notamment l'exploitation de cet outil en manipulation linguistique de corpus textuels.

\section{Une méthode linguistique de structuration textuelle}

L'examen des systèmes de question-réponse existants nous a donc amené à élaborer une stratégie bien différente, centrée sur une analyse linguistique des énoncés. Cette démarche s'inscrit toutefois dans la tradition du domaine, puisque la définition de l'information considérée appartient à une \textbf{perspective} syntaxico-sémantique et lexico-sémantique. Il s'agit en effet d'identifier les unités lexicales porteuses de sens, considérées comme l'information élémentaire, les relations (syntaxiques) entre ces éléments, ainsi que le sens lexical des lexèmes en contexte. D'autre part, les variations dans la formulation d'une même information sont classiquement compensées par une expansion de l'énoncé, où synonymes, hypéronymes, holonymes et autres dérivés morphologiques interviennent à plaisir.

L'expansion d'énoncé pratiquée généralement dans le domaine de question-réponse s'applique à la requête. Le principe en est simple: à chaque unité lexicale considérée comme significative est associée une liste de mots qui lui sont jugés équivalents, sous forme de synonymes, d'hyponymes et hypéronymes, de dérivés, etc. Ces listes d'expansions sont utilisées disjonctivement au lexème original lorsque la requête est proposée à un moteur de recherche. Mais si une telle expansion permet effectivement de résoudre dans un grand nombre de cas les problèmes de formulation, seule la maîtrise du sens de l'énoncé à expanser permet de sélectionner les reformulations qui conviennent dans le contexte courant. La figure \ref{blindExp} page \pageref{blindExp} distingue les expansions correctes (A) des expansions erronées (B), susceptibles d'apporter des réponses inadéquates. 

\begin{center}
\begin{figure}[ht]
\centering
\begin{tabular}{p{2.5cm}p{2.5cm}p{1cm}p{2.5cm}}
\multicolumn{4}{l}{Question: «~De quel \textbf{chef} Domitien est-il le \textbf{successeur}?~»}\\
  &           &  & \\
A & général   &  & héritier \\
  & empereur  &  & succéder \\
  & \dots     &  & \dots \\
  &           &  & \\
B & cuisinier &  & remplaçant \\
  & cheveu    &  & succédané \\
  & \dots     &  & \dots \\
  &           &  & \\
\multicolumn{4}{l}{Réponse:}\\
\multicolumn{4}{p{9cm}}{Second fils de Vespasien, Domitien succéda à l'empereur Titus et poursuivit la remise en ordre de l'État.}
\end{tabular}
\caption{Exemple d'expansion d'énoncé: le problème du sens}
\label{blindExp}
\end{figure} 
\end{center}

Une sélection du sens des unités lexicales de la requête est donc nécessaire pour que l'expansion puisse être effectuée en fonction du sens original, afin de limiter le bruit. Cependant, on sait depuis \cite{Weaver49} toute l'importance que prend le contexte -- et même son contexte syntaxique \cite{Reifler55} -- dans le choix du sens lexical d'un mot dans une phrase (désambiguïsation sémantique lexicale). De plus, le simple bon sens permet de constater que les questions que l'on peut poser au système sont généralement plus courtes que des phrases rédigées dans un document. Le contexte y est donc moins important que dans les textes interrogés. Par ailleurs, les grammaires syntaxiques des outils d'analyse existants fonctionnent habituellement moins bien sur des phrases interrogatives que sur des énoncés affirmatifs. Il est donc moins efficace de traiter la sémantique lexicale d'un lexème s'il apparaît dans une requête que s'il figure dans la base textuelle elle-même.


Dès lors, une démarche qui applique prioritairement une analyse syntaxico-sémantique aux documents plutôt qu'à la requête s'est imposée. Ce choix est d'autant plus opportun qu'il s'intègre pleinement à l'indexation du contenu des documents, indispensable lors de la phase de recherche, et qui consiste à recenser le contenu des documents. De plus, une telle approche centrée sur les documents présente comme avantage pratique de distinguer nettement et chronologiquement l'analyse et l'expansion des énoncés, et la phase d'interrogation. De la sorte, les traitements les plus lourds sont appliqués préalablement, et l'interrogation de la structure de l'information se fait de manière presque instantanée.

Le système comporte donc deux niveaux de fonctionnement. Le premier consiste à analyser les documents d'un point de vue morphologique, syntaxique et sémantique, puis à leur appliquer une expansion à l'aide d'informations provenant de ressources lexico-sémantiques (adjonction d'\textit{enrichissements}), et à stocker les résultats dans des index constituant une structure de toute l'information textuelle \cite{RouxJacquemin02}. La seconde étape a pour objet l'interrogation de cette structure à l'aide de questions ordinaires.

\section{Outils d'analyse textuelle}

Comme nous l'avons indiqué plus haut, divers outils d'analyse interviennent dans l'identification des éléments d'information présents dans les documents. Il s'agit d'identifier les éléments eux-mêmes, c'est-à-dire les mots significatifs, au travers d'une analyse morphologique; ensuite, les relations entre ces mots grâce à l'analyse syntaxique; enfin, la désambiguïsation sémantique permet de connaître la signification des mots dans leur contexte d'apparition. Voici une rapide description de ces outils.

\subsection{L'analyseur morphologique}

L'analyseur morphologique NTM (\textit{Normalizer, Tokenizer, Morphological analyzer}) que nous avons utilisé est un système de transducteurs à états finis développé au Centre de Recherche européen de Xerox (XRCE) \cite{ait98}. Ce système prend en entrée n'importe quelle chaîne de caractères en français et y applique des traitements de normalisation, de segmentation s'il s'agit de plusieurs unités lexicales, et propose les différentes analyses morphologiques possibles pour chacun des segments identifiés. La figure \ref{anaMorph} page \pageref{anaMorph} permet d'identifier les traitements appliqués par NTM à une phrase proposée en entrée. Il permet d'obtenir une version normalisée de chaque unité lexicale, son lemme et les informations morphologiques qui y sont associées sous la forme de traits attachés à la forme de départ.

\begin{center}
\begin{figure}[htb]
\centering
\begin{tabular}{l|l|l|l}
\multicolumn{4}{l}{\textbf{Son} deuxième fils [\dots]}\\
\multicolumn{4}{l}{}\\
\multicolumn{1}{c|}{Mot du texte}&\multicolumn{1}{c|}{lemme}&\multicolumn{1}{c|}{Analyse morphologique}&\multicolumn{1}{c}{Traits ajoutés}\\
\hline
\texttt{son}   & \texttt{son}   & \texttt{+PP3S+InvGen+SG+Poss} &\\
\texttt{son}   & \texttt{son}   & \texttt{+Masc+SG+Noun+}   & \textbf{+SOM+AGR}\\
\multicolumn{4}{l}{~}
\end{tabular}
\\
\begin{tabular}{l@{}lll@{}l}
\texttt{PP3S}   & ~Pronom personnel 3ème sg & ~ & \texttt{Masc} & ~Masculin\\
\texttt{InvGen} & ~Invariable en genre      & ~ & \texttt{SG}   & ~Singulier\\
\texttt{SG}     & ~Singulier                & ~ & \texttt{Noun} & ~Nom\\
\texttt{Poss}   & ~Possessif                & ~ & \textbf{SOM}  & ~Relatif au corps\\
                &                           & ~ & \textbf{AGR}  & ~Agriculture\\
\end{tabular}
\caption{Exemple d'analyse morphologique par NTM}
\label{anaMorph}
\end{figure}
\end{center}

Cet analyseur présente également une qualité qui a pu être exploitée avec succès. En effet, sa conception sous forme de transducteurs, et la présentation de ses résultats sous forme de traits attachés aux unités lexicales, permettent d'ajouter aisément certaines informations lexicales qui peuvent être utiles pour les traitements ultérieurs. Ainsi, on peut voir dans l'exemple que nous avons ajouté aux lexiques existants des informations sémantiques extraites d'un dictionnaire, qui seront utilisées ultérieurement lors de la phase de désambiguïsation. Dans l'intervalle, cette information subsiste attachée aux unités lexicales, mais elle reste virtuelle dans la mesure où elle n'intervient ni dans la désambiguïsation catégorielle, ni dans l'analyse syntaxique.

\subsection{L'analyseur syntaxique}

L'analyseur syntaxique XIP (\textit{Xerox Incremental Parser}) \cite{Roux99} est un moteur d'analyse syntaxique basé sur des grammaires de réécritures incrémentales. Il permet d'effectuer le cas échéant une désambiguïsation catégorielle d'énoncés étiquetés morphologiquement mais non désambiguïsés. Il propose surtout une analyse syntaxique de surface robuste de ces énoncés sous forme de dépendances entre des n{\oe}uds représentés sous la forme des unités lexicales équivalant à la tête des syntagmes minimaux (\textit{chunks}) concernés. Une représentation en arbre de chaque phrase, ainsi qu'un découpage en syntagmes minimaux sont également proposés, mais ils ne sont pas utilisés ici.

\begin{center}
\begin{figure}[htb]
\centering
\begin{tabular}{lll}
\multicolumn{3}{l}{Énoncé~: «~Il reconstruisit Rome ruinée par les incendies.~»}\\
\multicolumn{3}{l}{}\\
\multicolumn{3}{l}{Extraction des dépendances:}\\
&\texttt{SUBJ(reconstruisit,Il)}            & 2e argument sujet du 1er argument\\
&\texttt{SUBJ(ruinée,incendies)}            &  \\
&\texttt{VMOD[INDIR](ruinée,par,incendies)} & 3e argument compl. agent 1er argument \\
&\texttt{VARG[DIR](reconstruisit,Rome)}     & 2e argument COD du 1er argument \\
&\texttt{NMOD[ADJ](Rome,ruinée)}            & 2e argument épithète du 1er argument \\
\end{tabular}
\caption{Exemple d'analyse syntaxique par XIP}
\label{anaSynt}
\end{figure}
\end{center}

La figure \ref{anaSynt} page \pageref{anaSynt} permet d'évaluer les possibilités de XIP et d'illustrer son mode de représentation des relations syntaxiques par dépendances. On peut également voir le travail de certains traits, exclusivement syntaxique ici, et portant sur la nature des dépendances (\texttt{DIR} et \texttt{INDIR} respectivement sur les dépendances \texttt{VMOD} et \texttt{VARG}, ainsi que \texttt{ADJ} sur \texttt{NMOD}). XIP applique des règles contextuelles qui permettent d'évaluer des n{\oe}uds et des traits portant sur des n{\oe}uds appartenant à un même contexte pour construire des syntagmes minimaux et des dépendances. Ce mode de fonctionnement est très intéressant car s'il permet de travailler sur des indications lexico-morphologiques pour la désambiguïsation catégorielle et des données lexico-syntaxiques pour la construction des dépendances syntaxiques, il n'y a pas d'obstacle à son utilisation dans une perspective lexico-sémantique.

\subsection{Le désambiguïsateur sémantique}

Le système de désambiguïsation sémantique présenté de le cadre de cette étude est une évolution de la méthode de \cite{brunAl01}, qui reposait sur l'exploitation de l'analyse syntaxico-sémantique d'un dictionnaire utilisé comme corpus sémantiquement étiqueté. Le présent système \cite{Jacquemin03} exploite l'information du \textit{Dictionnaire des verbes français} \cite{dubois-duboischarlier97} et de son complément des autres catégories grammaticales (ces deux dictionnaires complémentaires seront désormais désignés sous le nom générique \textit{Dubois}). Ces dictionnaires répartissent l'information fournie non par unité lexicale, mais par sens de chaque unité lexicale. De la sorte, chaque information fournie par le dictionnaire est discriminante pour le sens concerné d'un mot donné. 

Le fonctionnement du désambiguïsateur se fait en deux temps: d'abord l'analyse du dictionnaire avec création de règles conditionnelles de désambiguïsation sémantique, basées sur un schéma syntaxique, et ensuite l'application de ces règles à des mots en contexte, sur base des contextes syntaxico-sémantiques fournis d'abord par les étiquettes sémantiques ajoutées à NTM, ensuite par les dépendances issues de l'analyse syntaxique de XIP.

L'information qui peut être extraite du dictionnaire régit le type de règles qui peuvent être construite. Dans le cas du Dubois, l'information peut être diverse et se présenter sous forme purement syntaxique (p.ex. «~Je bois~» \textit{vs} «~Je bois de l'eau~» avec l'indication de transitivité), syntaxico-sémantique (p.ex. «~embrasser quelque chose~» \textit{vs} «~embrasser quelqu'un~» avec sous-catégorisation du complément direct), lexico-syntaxique avec l'analyse des exemples et la conservation des relations impliquant le mot considéré comme autant de schémas typiques («~le général remporte la victoire~» implique la dépendance \texttt{VARG[DIR](remporter,victoire)}, avec le mot victoire comme complément direct de remporter) ou sémantico-syntaxique (généralisation de la dépendance extraite d'un exemple grâce aux traits sémantiques correspondant à une unité lexicale: \texttt{VARG[DIR](remporter,[MIL])}, où le trait \texttt{MIL} pour militaire est le trait sémantique de \textit{victoire}, qu'il remplace).

Comme les règles de désambiguïsation doivent répondre au contexte syntaxique, comme le stipule \cite{Reifler55}, et qu'elles sont conditionnelles -- puisque la conformité d'un contexte à une information issue d'un dictionnaire implique la sélection du sens correspondant -- elles répondent à toutes les conditions pour en faire une grammaire dans XIP. C'est donc à cette syntaxe que les règles de désambiguïsation doivent correspondre, ce qui évitera la création d'un moteur d'application des règles particulier. Les résultats d'une désambiguïsation sémantique apparaîtront donc comme des dépendances extraites par XIP ou comme des traits sur des dépendances ou des n{\oe}uds de XIP. La figure \ref{disambRule} page \pageref{disambRule} permet de comprendre le mode de construction de règles de désambiguïsation sémantique à partir de l'information contenue dans le dictionnaire. Les données syntaxiques ou lexicales sont formalisées sous la forme d'une dépendance XIP et les données sémantiques sous la forme de traits sur les n{\oe}uds.

\begin{center}
\begin{figure}[ht]
\centering
\begin{minipage}{11cm}
Exemple extrait du Dubois pour \textit{«~remporter~»} au sens \textit{03 gagner}:\\
\begin{center}«~Le général remporte la victoire~».\\\end{center}
\vspace{0.5cm}
Dépendances extraites de l'exemple:\\
$\mathtt{SUBJ(remporter,g\acute{e}n\acute{e}ral)}$\\
$\mathtt{\mathbf{VARG[DIR](remporter,victoire)}}$\\
\begin{description}
\item[$\bullet$] Construction d'une règle lexico-syntaxique de désambiguïsation:
\item[$\quad$] $\mathtt{remporter:\ VARG[DIR](}\mathit{remporter}\mathtt{,\mathbf{victoire})}\\ \Rightarrow\ \mathrm{remporter\ 03\ }$«~gagner~»
\item[$\longrightarrow$] apparition de victoire comme complément direct de remporter implique le sens 03 «~gagner~»\\
\end{description}
\begin{description}
\item[$\bullet$] Construction de la règle sémantico-syntaxique correspondante:
\item[$\quad$] victoire $\rightarrow$ trait sémantique: \texttt{MIL} (militaire)
\item[$\quad$] $\mathtt{remporter:\ VARG[DIR](}\mathit{remporter}\mathtt{,\mathbf{MIL})}\\ \Rightarrow\ \mathrm{remporter\ 03\ }$«~gagner~»
\item[$\longrightarrow$] apparition d'un mot comportant le trait \texttt{MIL} (militaire) comme complément direct de remporter implique le sens 03 «~gagner~»
\end{description}
\end{minipage}
\caption{Exemple de construction des règles de désambiguïsation}
\label{disambRule}
\end{figure} 
\end{center}

L'application des règles se fait au travers de l'analyse des énoncés par XIP. La mise en correspondance de l'analyse syntaxique de XIP avec le schéma syntaxico-sémantique d'une règle de désambiguïsation implique l'application de la règle, ou la sélection du sens correspondant à cette règle. Le sens sélectionné est indiqué sous la forme d'un trait associé avec l'unité lexicale considérée. Les informations sémantiques ajoutées par NTM servent à l'application des règles impliquant des indications sémantiques.

\section{Adaptation des ressources lexicales}

On a déjà pu voir que les données lexicales étaient capitales pour l'approche proposée ici. Elles le sont non seulement dans la perspective de l'analyse sémantique, où l'information syntaxico-sémantique distribuée par sens des entrées est prépondérante, mais aussi dans une optique d'expansion d'énoncés. En effet, cette expansion est effectuée par remplacement d'unités lexicales originales par d'autres, qui peuvent leur être substituées avec un minimum de modifications de sens. Ce sont donc deux types de modifications lexicales qui sont réalisées: la synonymie, et la dérivation morphologique. Dans une certaine mesure, le dictionnaire Dubois est à même de fournir les indications permettant de procéder à ces transformations.

En effet, un des champs informationnels de ce dictionnaire de référence fournit des synonymes, tandis qu'un autre procure des indications relatives à la dérivation. Toutefois, les synonymes sont invariablement au nombre de deux, ce qui est généralement insuffisant pour couvrir l'ensemble des transformations synonymiques possibles. D'autre part, les indications de dérivations se basent sur une racine et des suffixes, qu'un système automatique est difficilement à même d'interpréter correctement. Dès lors, d'autres ressources et outils doivent être exploités pour combler les lacunes du Dubois.

\subsection{Adjonction et répartition de synonymes}

Pour ajouter une information synonymique au Dubois, nous avons utilisé trois ressources lexico-sémantiques: EuroWordNet \cite{Vossen98,Catherin99}, le dictionnaire des synonymes de  \cite{bailly47}, et un dictionnaire multilingue utilisé comme outil chez Memodata. Tous fournissent des synonymes, mais leur répartition par sens, quand elle existe, ne correspond pas à celle du Dubois. Il a donc fallu les redistribuer. Nous avons élaboré une méthode qui le fait automatiquement, décrite ici \cite{Jacquemin04b}.

Cette procédure établit pour chaque entrée de chaque dictionnaire la liste des synonymes sans faire de distinction entre les sens différents que cette entrée peut avoir. Ensuite, à chaque synonyme proposé, elle associe toutes les étiquette sémantiques qui lui sont attachées dans le dictionnaire Dubois. Puis, pour l'entrée considérée, chaque sens du Dubois est considéré successivement: lorsqu'une des étiquettes sémantiques du synonyme proposé est identique à celle du sens courant de l'entrée, il est considéré comme un synonyme valable pour ce sens et ajouté au champ de synonymie du Dubois. La même opération est effectuée pour chaque entrée de chaque dictionnaire de synonymes, puis les doublons sont éliminés. La figure \ref{syno} page \pageref{syno} illustre la procédure suivie.

\begin{center}
\begin{figure}[ht]
\centering
\begin{tabular}{llp{1cm}ll}
\multicolumn{5}{l}{ravir (sens n°2, «~voler~»)}\\
\multicolumn{5}{l}{~}\\
Synonymes proposés: & \textbf{enlever} & & étiquette sémantique & \textbf{SOC} / LOC / TEX\dots\\
                    & charmer & &                      & PSY / OCC \\
		    & \dots   & &                      & \dots \\
\multicolumn{5}{l}{~}\\
\multicolumn{5}{l}{Étiquette sémantique de ravir (2): \textbf{SOC}}\\
\multicolumn{5}{l}{$\Rightarrow$ synonyme ajouté: \textbf{enlever}}
\end{tabular}
\caption{Répartition des synonymes par sens du mot original}
\label{syno}
\end{figure} 
\end{center}

\subsection{Génération de dérivés}

D'autre part, l'information contenue dans le Dubois ne permet pas d'effectuer automatiquement la génération des formes dérivées à partir d'une vedette du dictionnaire. Par contre, cette information peut se révéler suffisante pour identifier une proposition de dérivation et confirmer sa validité. L'outil de dérivation morphologique proposé par \cite{Gaussier99} peut dès lors être utile puisqu'il permet de générer, pour un mot proposé, un très grand nombre de candidats dérivés qui sont également des lexèmes attestés dans le lexique, à condition de lui laisser un maximum de latitude en diminuant au maximum les contraintes de génération. Les données de dérivation indiquées dans les champs correspondants du Dubois permettent ensuite, par identification du suffixe et de certaines caractéristiques de la racine, de ne conserver pour chaque sens que les dérivés prescrits par le dictionnaire.


\section{Construction de la structure informationnelle}

Comme on l'a vu plus haut, la structure de l'information est constituée d'index comprenant l'ensemble des données contenues dans les dictionnaires, et permettant d'avoir accès directement à la bribe de texte considérée comme intéressante dans la base documentaire. Cette structure est constituée d'abord du résultat de l'analyse des textes, c'est-à-dire des unités lexicales identifiées lors de l'analyse morphologique, ainsi que des traits morphologiques qui y sont associés, puis des relations syntaxiques entre ces lexèmes, ainsi que des traits syntaxiques associés soit aux dépendances, soit aux unités lexicales, et enfin des traits sémantiques dénotés lors de la désambiguïsation sémantique (numéro de sens, indications sémantiques propres à ce sens dans le Dubois), qui portent uniquement sur les unités lexicales. La figure \ref{struct} page \pageref{struct} montre comment la structure informationnelle est construite à partir d'un énoncé et de son analyse: les traits sont représentés entre crochets, les dépendances en majuscules et les unités lexicales en minuscules.

\begin{center}
\begin{figure}[ht]
\begin{center}
\begin{minipage}{12cm}
\begin{center}
«~\dots Domitien succéda à l'empereur Titus\dots~»
\end{center}
\vspace{0cm}
\begin{tabbing}
Retrait\= suite \kill
 \> \texttt{SUBJ(succéda[sn=1],Domitien[proper])}\\
 \> \texttt{VARG[INDIR](succéda[sn=1,],à,empereur[humain,])}\\
 \> \texttt{NN(empereur,Titus[proper])}\\
\end{tabbing} 
\end{minipage}
\end{center}
\caption{Construction du «~squelette~» de la structure informationnelle}
\label{struct}
\end{figure} 
\end{center}

Dans un deuxième temps, la structure informationnelle est enrichie par l'expansion des énoncés. Les synonymes sont ajoutés disjonctivement aux dépendances dans lesquelles apparaissent les unités lexicales originales. Par contre, les formes dérivées ne peuvent être placées de même dans la structure informationnelle, car elles appartiennent le plus souvent à une catégorie grammaticale différente de celle du lexème original dont elles sont dérivées, et ne présentent pas une construction syntaxique similaire. Pour conserver une signification aussi proche que possible de l'énoncé original, il s'agit donc de reformuler l'énoncé pour qu'il intègre la forme dérivée. Pour cela, nous avons étudié le processus de dérivation: pour chaque catégorie grammaticale originale, pour chaque catégorie grammaticale dérivée, pour chaque type suffixal de dérivation, nous avons sélectionné au hasard trois exemples de dérivation dans le Dubois, et nous avons observé de quelle manière on peut remplacer l'original par le dérivé dans vingt contextes réels obtenus sur le Web. À partir de là se sont dégagés des patrons de dérivation qui permettent à partir d'un contexte sémantico-lexico-syntaxique, d'identifier le schéma syntaxique d'apparition de l'original et d'inférer un schéma syntaxique de dérivation. La figure \ref{structExp} page \pageref{structExp} permet de comprendre de quelle manière les différents enrichissements sont ajoutés à la structure originale pour constituer des expansions des énoncés originaux, que ce soit par synonymie ou dérivation.

\begin{center}
\begin{figure}[ht]
\begin{center}
\begin{minipage}{12cm}
\begin{center}
«~\dots Domitien succéda à l'empereur Titus\dots~»
\end{center}
\vspace{0cm}
\begin{tabbing}
Retrait\= suite \kill
Résultats avant expansion:\\
 \> \texttt{SUBJ(succéda[sn=1],Domitien[proper])}\\
 \> \texttt{VARG[INDIR](succéda[sn=1,],à,empereur[humain,])}\\
 \> \texttt{NN(empereur,Titus[proper])}\\
\end{tabbing} 
\begin{tabbing}
Retrait\= suite \kill
Structure de l'énoncé avec expantion:\\
 \> \texttt{SUBJ(succéda/remplacer,Domitien)}  \\
 \> \texttt{VARG[INDIR](succéda,à,empereur)}  \\
 \> \texttt{VARG[DIR](remplacer,empereur/chef/[\dots])}  \\
 \> \texttt{NN(empereur/chef/souverain/dots,Titus)}  \\
 \> \vspace{0.1cm}\\
 \> \texttt{NMOD[INDIR](successeur,de,empereur/chef/souverain/\dots)}  \\
 \> \texttt{NMOD(Domitien,successeur)} \\
\end{tabbing}
\vspace{0.3cm}
\end{minipage}
\end{center}
\caption{Construction de la structure informationnelle avec expansion}
\label{structExp}
\end{figure} 
\end{center}

\section{Interrogation de la structure}

L'interrogation de la structure informationnelle peut être effectuée de nombreuses façons, dans la mesure où il suffit d'effectuer une recherche sur un ou plusieurs éléments stockés dans les index pour obtenir instantanément les énoncés d'apparition de ces éléments. Dans le cadre de l'application de question-réponse, c'est à une question en langue naturelle que le système doit apporter une réponse \cite{Jacquemin04a}. 
L'information contenue dans la question doit donc être convertie dans un format similaire à celui de la structure informationnelle, c'est à dire dans une structure locale similaire. Toutefois, comme le contexte d'une question est insuffisant pour effectuer une analyse sémantique congrue, cette structure locale est légère, c'est-à-dire qu'elle est limitée aux analyses morphologique et syntaxique, excluant donc la désambiguïsation sémantique et la phase d'expansion.

Certaines particularités doivent pourtant être signalées dans la conception de cette structure légère de la question. En effet, une grammaire particulière est mise en {\oe}uvre dans l'analyse de la question, qui permet deux adaptations de la structure. La première réside dans la relation \texttt{FOCUS}, qui permet de caractériser l'objet de la question, et donc la réponse attendue. Il s'agit d'une dépendance de marquage, qui identifie l'unité lexicale la plus significative de l'interrogation, c'est-à-dire la tête du groupe nominal lorsque l'interrogatif est un adjectif («~Qui est le beau-père de Galère?~» \texttt{FOCUS(beau-père[PAR])}) ou l''interrogatif lui-même si c'est un pronom («~Qui combattit les Parthes?~» \texttt{FOCUS(qui[humain])}). Elle permet d'identifier les traits sémantiques de l'objet de la question, et donc d'identifier la réponse dans les documents lorsque les autres éléments de la question se trouvent dans un énoncé de la base textuelle. Cette dépendance n'existant pas dans les documents -- ni dans la structure informationnelle -- étant donné ce qu'elle représente, elle devra ensuite être transmise comme un trait à l'intérieur de la structure légère, et supprimée comme dépendance cette structure légère.

La seconde adaptation de la structure locale à la question réside dans la suppression de toutes les informations ne relevant que du caractère interrogatif de cette question. Ainsi, l'interrogatif sera supprimé pour ne conserver, au sein des dépendances qui le contiennent, que les traits sémantiques qui lui sont propres et, le cas échéant, le trait \texttt{FOCUS}. Les dépendances purement fonctionnelles disparaissent également (dues au fonctionnement interne de XIP ou mettant en {\oe}uvre des mots-outils, des auxiliaires ou semi-auxiliaires), car elles ne sont pas porteuses d'information pertinente dans le cadre de cette application. La dépendance FOCUS sera éliminée de même, mais le trait subsiste dans les dépendances où doit apparaître le lexème sur lequel porte cette dépendance.

\begin{center}
\begin{figure}
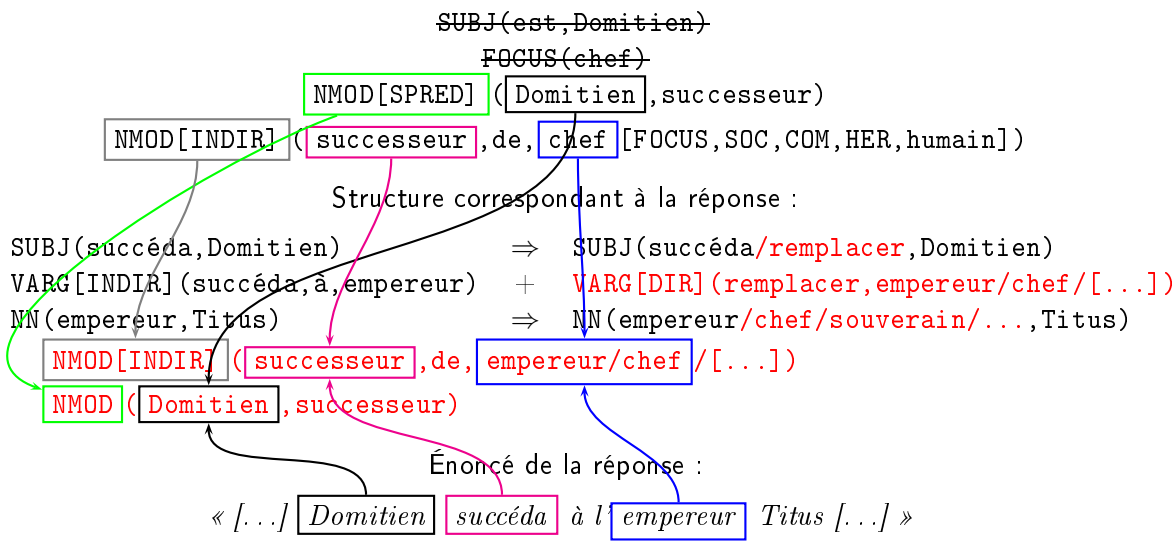

\begin{center}
\textit{«~De quel chef Domitien est-il le successeur?~»}\\
\vspace{0.3cm}
\textsf{Structure légère de la question:}\\
\vspace{0.2cm}\texttt{
\sout{SUBJ(est,Domitien)}\\
\sout{FOCUS(chef)}\\
\rnode{dep1A}{\psframebox[linecolor=green]{NMOD[SPRED]}}(\rnode{domiA}{\psframebox[linecolor=black]{Domitien}},successeur)\\
\rnode{dep2A}{\psframebox[linecolor=gray]{NMOD[INDIR]}}(\rnode{succA}{\psframebox[linecolor=magenta]{successeur}},de,\rnode{chefA}{\psframebox[linecolor=blue]{chef}}[FOCUS,SOC,COM,HER,humain])}\\
\vspace{0.3cm}
\textsf{Structure correspondant à la réponse:}\\
\vspace{0.2cm}
\begin{tabular}{llcl}
\multicolumn{2}{l}{\texttt{SUBJ(succéda,Domitien)}}          & $\Rightarrow$ & \texttt{SUBJ(succéda{\red /remplacer},Domitien)}  \\
\multicolumn{2}{l}{\texttt{VARG[INDIR](succéda,à,empereur)}} & +             &\texttt{{\red VARG[DIR](remplacer,empereur/chef/[\dots])}}\\
\multicolumn{2}{l}{\texttt{NN(empereur,Titus)}}              & $\Rightarrow$ & \texttt{NN(empereur{\red /chef/souverain/\dots},Titus)}  \\
& \multicolumn{3}{l}{\texttt{{\red\rnode{dep2B}{\psframebox[linecolor=gray]{NMOD[INDIR]}}(\rnode{succB}{\psframebox[linecolor=magenta]{successeur}},de,\rnode{chefB}{\psframebox[linecolor=blue]{empereur/chef}}/[\dots])}}}\\
& \multicolumn{3}{l}{\texttt{{\red\rnode{dep1B}{\psframebox[linecolor=green]{NMOD}}(\rnode{domiB}{\psframebox[linecolor=black]{Domitien}},successeur)}}}\\
\end{tabular}\\
\vspace{0.3cm}
\textsf{Énoncé de la réponse~:}\\
\vspace{0.2cm}
\textit{«~[\dots] \rnode{domiC}{\psframebox[linecolor=black]{Domitien}} \rnode{succC}{\psframebox[linecolor=magenta]{succéda}} à l'\rnode{chefC}{\psframebox[linecolor=blue]{empereur}} Titus [\dots]~»}
\nccurve[linecolor=green,angleA=200,angleB=160]{->}{dep1A}{dep1B}
\nccurve[linecolor=gray,angleA=270,angleB=90]{->}{dep2A}{dep2B}
\nccurve[linecolor=black,angleA=270,angleB=90]{->}{domiA}{domiB}
\nccurve[linecolor=black,angleA=90,angleB=270]{->}{domiC}{domiB}
\nccurve[linecolor=magenta,angleA=270,angleB=90]{->}{succA}{succB}
\nccurve[linecolor=magenta,angleA=90,angleB=270]{->}{succC}{succB}
\nccurve[linecolor=blue,angleA=270,angleB=90]{->}{chefA}{chefB}
\nccurve[linecolor=blue,angleA=90,angleB=270]{->}{chefC}{chefB}
\end{center}
\caption{Exemple d'interrogation de la structure informationnelle avec son expansion}
\label{interro} 
\end{figure} 
\end{center}

La recherche d'une réponse revient donc à mettre en correspondance la structure légère de la question, débarrassée de l'information propre à une interrogation, et des bribes de texte au travers de la structure informationnelle. Lorsqu'une information concordante à la structure légère est trouvée au sein de la même phrase dans la structure de l'information, cette phrase est considérée comme une réponse pertinente à la question. Bien entendu, la réponse est considérée comme plus pertinente si une plus grande partie de l'information qui concorde est originale dans le texte, et moins pertinente à mesure que ces éléments concordants sont issus d'une expansion. La figure \ref{interro} page \pageref{interro} illustre bien la mise en concordance d'une question avec sa réponse au travers de deux structures, l'une légère et purifiée, l'autre complète et enrichie d'expansions.

\section{Conclusion}

Nous avons présenté un système généraliste d'interrogation d'une base documentaire textuelle en langue naturelle. Ce système s'appuie sur des bases théoriques et sur des constatations pratiques pour proposer une méthode originale de structuration de l'information dans une base textuelle avec expansion des documents plutôt que des requêtes. L'ensemble des analyses et des enrichissements ont été effectués par des analyseurs linguistiques et les choix ont été réalisés suivant des indices contextuels et symbolistes issus de grammaires décrivant la langue.

Une analyse de ce système n'a pu être présentée ici faute de place. On peut en trouver le détail dans \cite{Jacquemin03}. Il montre la validité de la méthode -- elle soutient la comparaison avec les meilleurs systèmes de sa catégorie dans la conférence TREC --, ainsi que certaines faiblesses, essentiellement liées à l'absence de résolution d'anaphores ou de hiérarchie sémantique. D'autre part, cette approche souffre, comme c'est habituel dans le domaine, de la représentation exclusivement lexicale de l'information, qui tient peu compte des mécanismes logiques. L'inférence, par exemple, n'est pas gérée actuellement, mais certaines approches statistique du lexique sont prometteuses à ce stade.

Par ailleurs, la présentation de ce système a été faite uniquement dans une optique de gestion de l'information. Cependant, il pourrait également se révéler un précieux outil d'étude de corpus écrit, dans la mesure où il peut être interrogé aisément et rapidement, que tous les niveaux d'information linguistique sont disponibles à tout moment et qu'ils peuvent être individualisés sans problème. Ainsi, on peut facilement mêler dans une même requête des exigences lexicales, morphologiques, syntaxiques, sémantiques, de cooccurrence, et obtenir l'ensemble des réponses pertinentes quel que soit le corpus désiré, puisque ce système est automatique et qu'il accepte du texte tout venant avec une robustesse inhabituelle. Une telle approche à dominante linguistique semble dès lors se justifier, même si des améliorations peuvent et doivent y être apportées.



\bibliographystyle{apalike-fr}
\bibliography{bj}


\end{document}